\documentclass[11pt,a4paper]{article}
\usepackage[hyperref]{naaclhlt2019}
\usepackage{times}
\usepackage{latexsym}
\usepackage{multirow}
\usepackage{graphicx}
\usepackage{epstopdf}
\usepackage{tabularx}
\usepackage{float}
\usepackage{subcaption}
\usepackage{cleveref}
\usepackage{adjustbox}

\usepackage{url}
\usepackage{amsmath}

\usepackage[utf8]{inputenc}
\usepackage{booktabs}

\usepackage{todonotes} 
\makeatletter
\newcommand*\iftodonotes{\if@todonotes@disabled\expandafter\@secondoftwo\else\expandafter\@firstoftwo\fi}  
\makeatother

\aclfinalcopy 

\title{Zero-Shot Cross-Language Relation Extraction\\as Multilingual Machine Comprehension}

\title{Extracting Unseen Relations in Unseen Languages}

\title{X-WikiRE: A Large, Multilingual Resource for Relation Extraction as Machine Comprehension}

\author{Mostafa Abdou, Cezar Sas, Rahul Aralikatte, Isabelle Augenstein \and Anders Søgaard\\
{\tt \{abdou, sas, rahul, augenstein, soegaard\} @ di.ku.dk}\\
University of Copenhagen
} 

\date{}

\begin{document}
\maketitle
\begin{abstract}

Although the vast majority of knowledge bases (\textbf{KBs}) are heavily biased towards English, Wikipedias do cover very different topics in different languages. Exploiting this, we introduce a new multilingual dataset (\textbf{X-WikiRE}), framing relation extraction as a multilingual machine reading problem. We show that by leveraging this resource it is possible to robustly transfer models cross-lingually and that multilingual support significantly improves (zero-shot) relation extraction, enabling the population of low-resourced \textbf{KBs} from their well-populated counterparts.

\end{abstract}

\section{Introduction}

It is a widely lamented fact that linguistic and encyclopedic resources are heavily biased towards English.
Even multilingual knowledge bases (KBs) such as Wikidata \citep{vwikidata} are predominantly English-based \cite{Kaffee:Simperl:18}. This means that coverage is higher for English, and that facts of interest to English-speaking communities are more likely included in a KB. This work introduces a novel multilingual dataset ({\bf X-WikiRE}) and explores techniques for automatically filling such language gaps by learning, from {\bf X-WikiRE}, to add facts in other languages. Finally, we show that multilingual sharing is beneficial for knowledge base completion across all languages, including English.

The task of identifying potential KB entries in running text -- i.e., relations that hold between two or more entities, is called {\em relation extraction} (RE). In the traditional, supervised setting \cite{bach2007review}, RE models are trained to identify a pre-specified set of relation types, which are observed during training. Models are meant to generalize to new entities, but {\em not} new {\em relations}. An alternative flavor is {\em open} RE \cite{fader2011identifying, yates2007textrunner}, which detects subject-verb-object triples and clusters semantically related verbs into  coarse-grained semantic relations. 

In this paper, we consider the middle ground, in which models are trained on a subset of pre-specified relations and applied to both seen and unseen entities, and unseen relations. The latter scenario is known as {\em zero-shot} RE \cite{rocktaschel2015injecting}.  

\begin{figure}[t]
\centering
\includegraphics[width=0.4\textwidth]{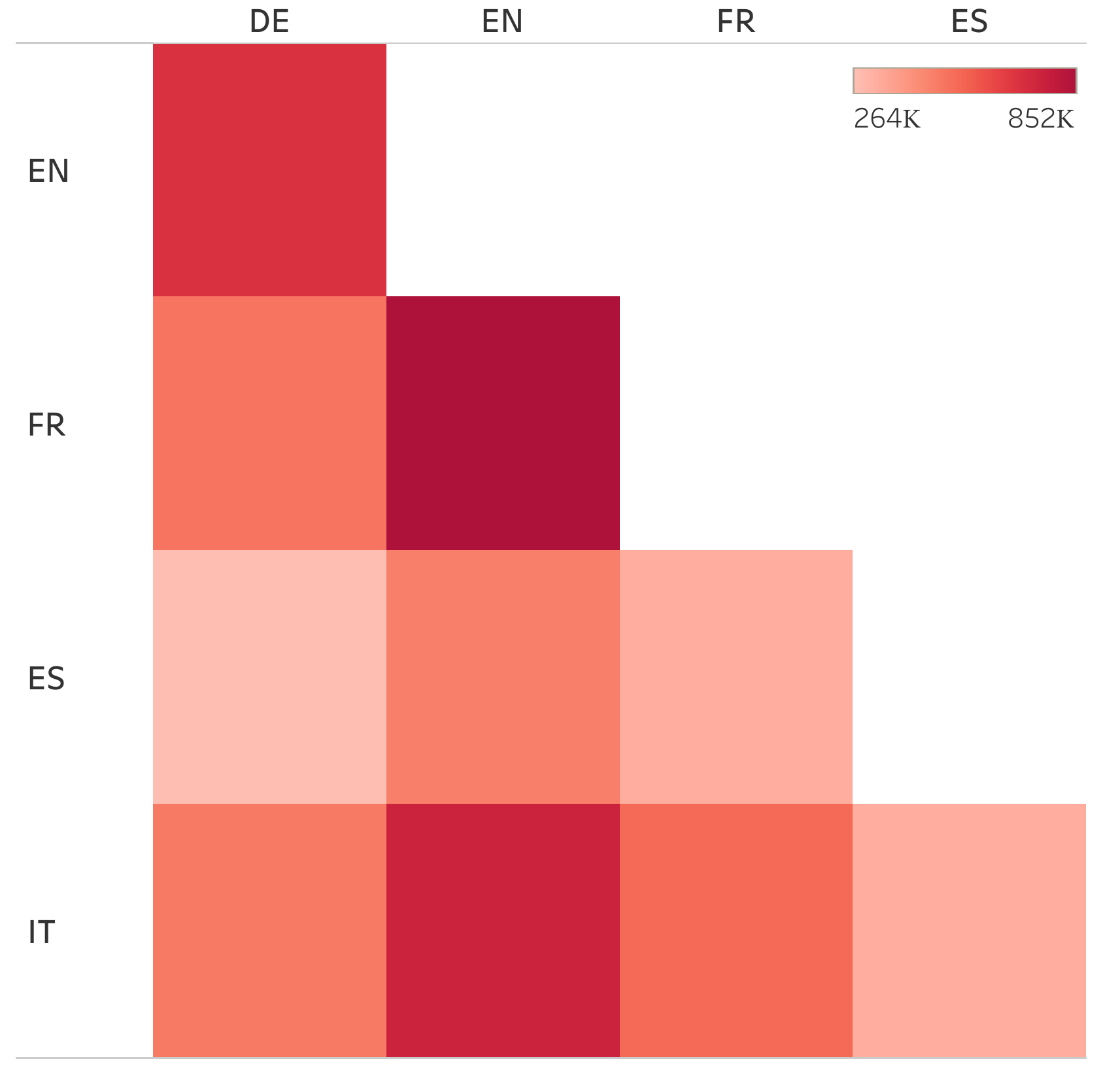}
\caption{The overlap of triples between languages.}
\label{fig:intersection_heatmap}
\end{figure}

\newcite{levy2017zero} present a reformulation of RE, where the task is framed as reading comprehension. In this formulation, each relation type (e.g. $author$, $occupation$) is mapped to at least one natural language question template (e.g. ``\textit{Who is the author of x?}"), where $x$ is filled with an entity (e.g. ``\textit{Inferno}"). The model is then tasked with finding an answer (``\textit{Dante Alighieri}'') to this question with respect to a given context. They show that this formulation of the problem both outperforms off-the-shelf RE systems in the typical RE setting and, in addition, enables generalization to unspecified and unseen types of relations. {\bf X-WikiRE} enables exploration of this reformulation of RE in a multilingual setting.

\begin{figure*}
\centering
\includegraphics[width=\textwidth]{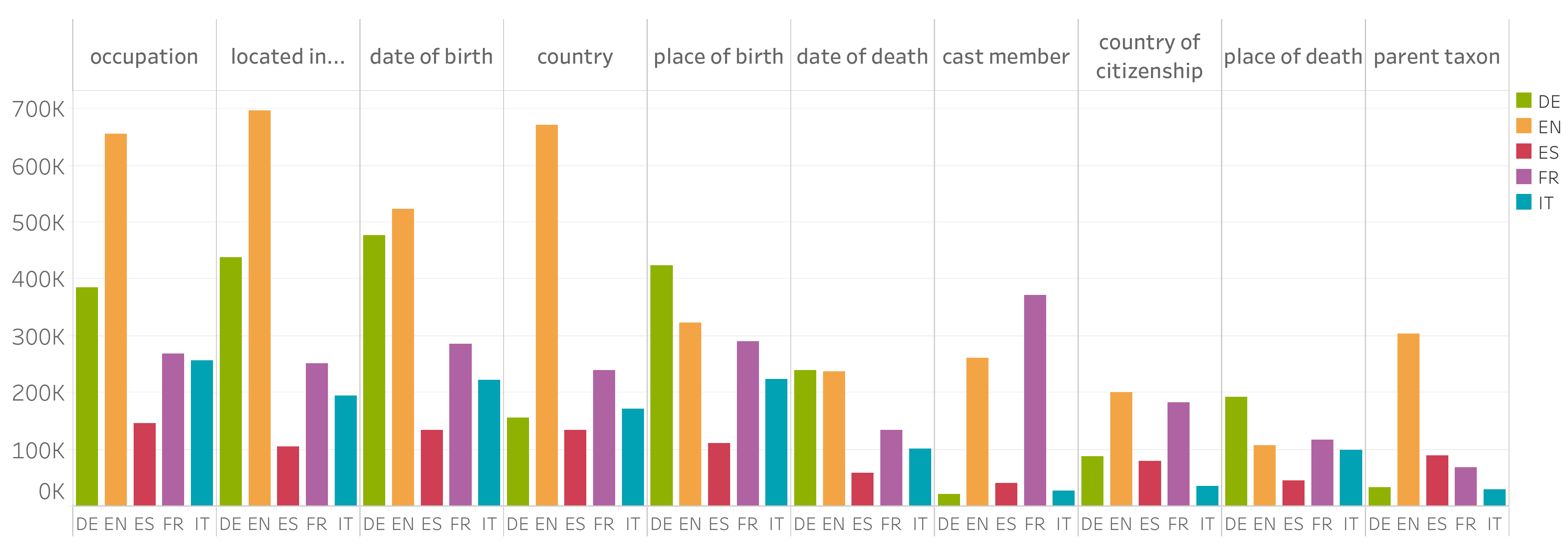}
\caption{The number of triples for the top 10 properties in each language.}
\label{fig:2}
\end{figure*}

\paragraph{Contributions} We introduce a new, large-scale multilingual dataset (\textbf{X-WikiRE}) of reading comprehension-based RE for English, German, French, Spanish, and Italian, facilitating research on multilingual methods for RE. Our dataset covers more languages (five) and is at least an order of magnitude larger than existing multilingual RE datasets, e.g., TAC 2016 \citep{ellis2015overview}, which covers three languages and consists of $\approx$ 90k examples. We also a) perform cross-lingual RE showing that models pretrained on one language can be effectively transferred to others with minimal in-language finetuning; b) leverage multilingual representations to train a model capable of simultaneously performing (zero-shot) RE in all five languages, rivaling or outperforming its monolingually trained counterparts in many cases while requiring far fewer parameters per language; c) obtain considerable improvements by employing a more carefully designed nil-aware machine comprehension model.

\begin{table*}[h!]
\fontsize{10}{10}\selectfont
\centering
\begin{adjustbox}{width=\textwidth}
\begin{tabular}{p{0.75cm}p{4cm}p{10cm}}
\toprule
Lang & Question                                                      & Context \& Answers                                                                                                                                                                                                                      \\ \midrule
DE       & In welchem land befindet man sich, wenn man \textbf{Amazonas} besucht? & Der Fluss Amazonas gab seinerseits dem Amazonasbecken sowie mehreren gleichnamigen Verwaltungseinheiten in \textbf{Brasilien}, \textbf{Venezuela}, \textbf{Kolumbien} \dots                                                                         \\ \midrule
EN       & What country is \textbf{Amazon} located in?                            & The Amazon proper runs mostly through \textbf{Brazil} and \textbf{Peru}, and is part of the border between \dots                                                                                                                             \\ \midrule
ES       & ¿En qué país se encuentra el \textbf{Amazonas}?                        & El río Amazonas es un río de América del Sur, que atraviesa \textbf{Perú}, \textbf{Colombia} y \textbf{Brasil}.                                                                                                                       \\ \midrule
FR       & Dans quel pays peux-tu trouver \textbf{Amazone}?                       & Le fleuve prend alors le nom d'Amazonas au \textbf{Pérou} et en \textbf{Colombie}, puis celui de rio Solimões en entrant au \textbf{Brésil} au \dots \\ \midrule
IT       & Di quale nazione fa parte il \textbf{Rio delle Amazzoni}?              & Il Rio delle Amazzoni è un fiume dell'America Meridionale che attraversa \textbf{Perù}, \textbf{Colombia} e \textbf{Brasile} \dots                                                                    \\ 
\bottomrule
\end{tabular}
\end{adjustbox}
\caption{Examples from our dataset of the same question-context pairs across all the languages with the correct answers highlighted in boldface.}
\label{table:qa_examples}
\end{table*}

\section{Background}
\label{sec:background}

\paragraph{Relation extraction}
We begin with a brief description of our terminology. Given raw text, relation extraction is the task of identifying instances of relations $relation(entity_1, entity_2)$. We refer to these instances of relation and entity pairs as \textit{triples}. Furthermore, throughout this work, we use the term \textit{property} interchangeably with relation.

A large part of previous work on relation extraction has been concerned with extracting relations between unseen entities for a pre-defined set of relations seen during training \citep{zelenko2003kernel, guodong2005exploring, miwa2016end}. For example, the instances \texttt{(Barack Obama, Hawaii)}, \texttt{(Niels Bohr, Copenhagen)}, and \texttt{(Jacques Brel, Schaerbeek)} of the relation $born\_{in}(x,y)$ would be seen during the training phase, and then the model would be expected to correctly identify other instances of the relation such as \texttt{(Jean-Paul Sartre, Paris)} in running text. This is useful in closed-domain settings where it is possible to pre-select a set of relations of interest. 
In an open-domain setting, however, we are interested in the far more difficult problem of extracting unseen relation types. Open RE methods \citep{yates2007textrunner,banko2007open,fader2011identifying} do not require relation-specific data, but treat different phrasings of the same relation as different relations and rely on a combination of syntactic features (e.g. dependency parses) and normalisation rules, and so have limited generalization capacity.

\paragraph{Zero-shot relation extraction} 
\label{para:zero-shot}

\newcite{levy2017zero} propose a novel approach towards achieving this generalization by transforming relations into natural language question templates. For instance, the relation $born\_{in}(x,y)$ can be expressed as ``\textit{Where was $x$ born?}" or ``\textit{In which place was $x$ born?}". Then, a reading comprehension model \cite{seo2016bidirectional,chen2017reading} can be trained on question, answer, and context examples where the $x$ slot is filled with an entity and the $y$ slot is either an answer if the answer is present in the context, or \texttt{NIL}.  The model is then able to extract relation instances (given expressions of the relations as questions) from raw text. To test this \textit{``harsh zero-shot''} setting of relation extraction, they build a dataset for RE as machine comprehension from WikiReading \citep{hewlett2016wikireading}, relying on alignments between Wikipedia pages and Wikidata KB triples. They show that their reading comprehension model is able to use linguistic cues to identify relation paraphrases and lexico-syntactic patterns of textual deviation from questions to answers, enabling it to identify instances of new relations. Similar work \citep{obamuyide2018zero} recently also showed that RE can be framed as natural language inference.

\section{X-WikiRE}
\textbf{X-WikiRE} is a multilingual reading comprehension-based relation extraction dataset. Each example in the dataset consists of a \textit{question}, a \textit{context}, and an \textit{answer}, where the question is a querified relation and the context may contain the answer or an indication that it is not present (\texttt{NIL}). Questions are obtained by transforming relations into question templates with slots where an entity is inserted. Within the RE framework described in Section \ref{sec:background}, $entity_1$ is filled into a slot in the question template and $entity_2$ is the answer. Each triple\footnote{Not to be confused with an example as an example contains an instantiation of a relation in the form of a question. Thus, the different question templates for each relation share the same id.} in the dataset can be identified uniquely across all languages. We construct \textbf{X-WikiRE} using the relevant parts of Wikidata and Wikipedia for each language. Wikidata is an open KB where the knowledge contained in each document is expressed as a set of \texttt{statements}, and each statement is a tuple \texttt{(property\_id, value\_id)} (e.g. statement \texttt{(P50, Q1067)} where P50 refers to $author$ and \texttt{Q1067} to ``\textit{Dante Alighieri}"). We perform data integration on Wikidata, as described by \newcite{hewlett2016wikireading}: for each entity in Wikipedia we take the corresponding Wikidata document, add the Wikipedia page text, and denormalize the statements. This consists of replacing the property and value ids of each statement in the document with the text label for values which are entities, and with the human readable form for numeric values (e.g. timestamps are converted to natural forms like ``\textit{25 May 1994}") obtaining a tuple $(property, entity)$.\footnote{We make the simplification of referring to all values as entities.}

\paragraph{Slot-filling data} To extract the contexts for each triple in our dataset we use the distant supervision method described by \newcite{levy2017zero}. For each Wikidata document belonging to a given $entity_1$ we take all the denormalized tuples $(property, entity_2)$ and extract the first sentence in the text containing both $entity_1$ and $entity_2$. Negatives (contexts without answers) are constructed by finding pairs of triples with common $entity_2$ type (to ensure they contain good distractors), swapping their context if $entity_2$ is not present in the context of the other triple. 

\paragraph{Querification} \newcite{levy2017zero} created 1192 question templates for 120 Wikidata properties. A template contains a placeholder for an entity \textit{x} (e.g. for property ``\textit{author}", some templates are ``\textit{Who wrote the novel x?}" and ``\textit{Who is the author of x?}"), which can be automatically filled in to create questions so that $question \approx template(property,x))$. For our multilingual dataset, we had these templates translated by human translators. The translators attempted to translate each of the original 1192 templates. If a template was difficult to translate, they were instructed to discard it. They were also instructed to create their own templates, paraphrasing the original ones when possible. This resulted in a varying number of templates for each of the properties across languages. In addition to the entity placeholder, some languages with richer morphology (Spanish, Italian, and German) required extra placeholders in the templates because of agreement phenomena (gender). We added a placeholder for definite articles, as well as one for gender-dependent filler words. The gender is automatically inferred from the Wikipedia page statistics and a few heuristics. Table \ref{table:qa_examples} shows the same example across five languages.

\paragraph{Dataset statistics} Table \ref{data:stats} shows the number of positive and negative triples and examples (i.e with and without consideration of the templates). 

As expected (due to the size of its Wikidata), English has the highest number of triples for most properties. However, as Figure \ref{fig:2} shows, there are properties where it has fewer triples than other languages (e.g. French has more triples for film related properties such as $cast\_member$ and $nominated\_for$). 
Figure \ref{fig:intersection_heatmap} shows the overlap in the number of  triples between different languages. While it can be seen that English, once again, has the highest overall overlap with the other languages, there are interesting deviations from this pattern where for certain properties other languages share a larger intersection (see Appendix \ref{appendix:multiwiki} for examples). 

\begin{table}[t!]
\fontsize{10}{10}\selectfont
\label{table:datasetstats}
\centering
\begin{tabular}{c | c c c c}
\toprule
Language & Pos  & Neg  & Pos*  & Neg* \\ \midrule
DE       & 2.5M & 545K & 11M  & 2.3M \\ 
EN       & 5.1M & 1M   & 64M  & 12M  \\ 
ES       & 1.2M & 211K & 5.5M & 1.1M \\ 
FR       & 2.3M & 867K & 18M  & 6.8M \\ 
IT       & 1.9M & 217K & 10M  & 1.2M \\ 
\bottomrule

\end{tabular}
\caption{The number of positive and negative triples for each language with (*) and without templates.}
\label{data:stats}
\end{table}

\begin{figure}[t!]
\centering
\includegraphics[width=0.4\textwidth]{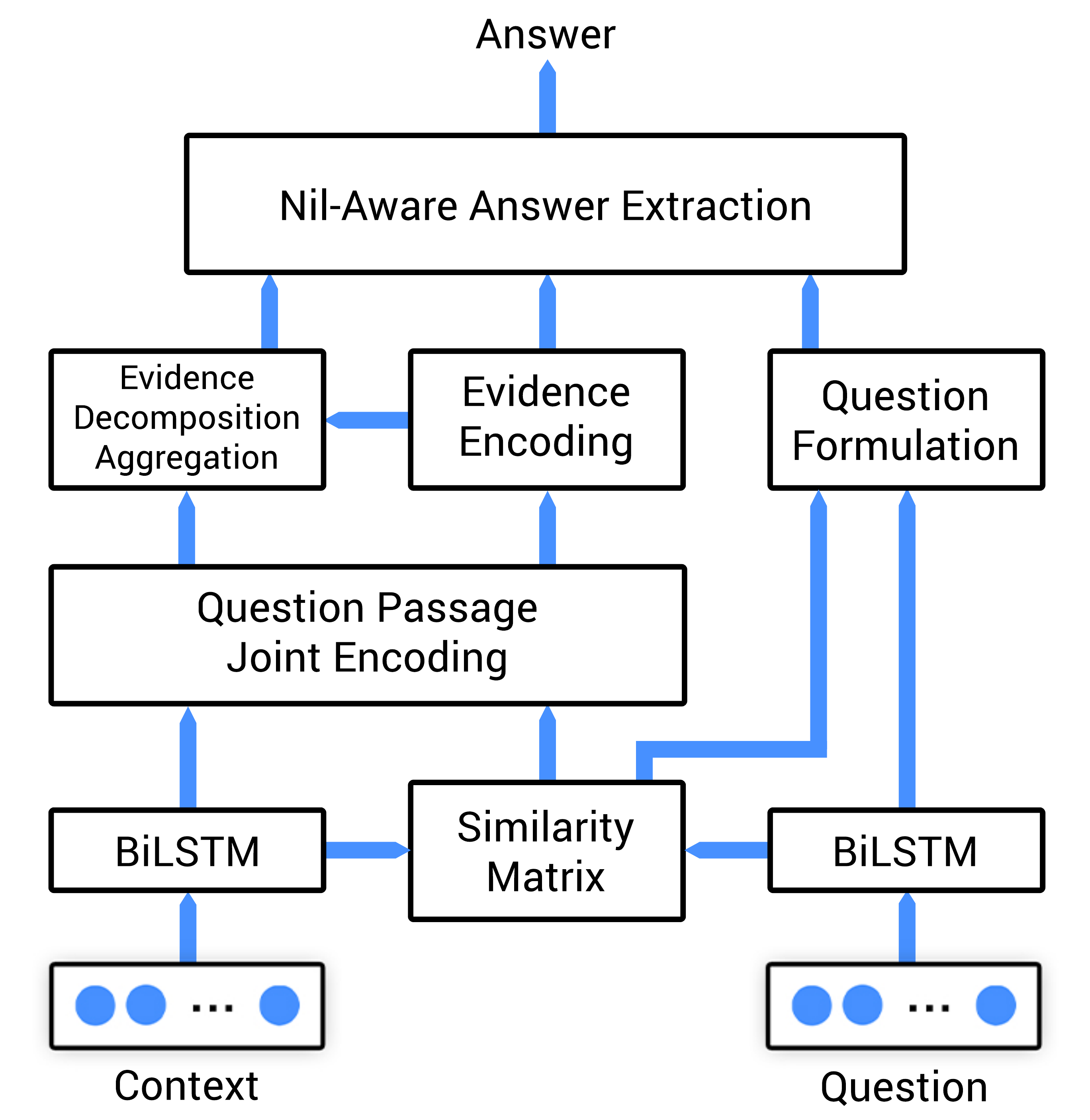}
\caption{An overview of Namanda's architecture.}
\label{fig:Namanda}
\end{figure}

\section{Method}
In our framework, a machine comprehension model sees a question-context pair and is tasked with selecting an answer span within the context, or indicating that the context does not contain an answer (returning \texttt{NIL}). This `nil-awareness' goes beyond the traditional reading comprehension setup where it is not required. It has, however, recently been incorporated into newer datasets \citep{trischler2016newsqa, rajpurkar2018know, saha2018duorc}. We employ the architecture described in \newcite{kundu2018namanda} as our standard reading comprehension model for all the experiments. This nil-aware answer extraction framework (NAMANDA) is briefly described below. In a set of initial trials (see Table \ref{table:bigtable}), we found that this model far outperformed the bias-augmented BiDAF model \cite{seo2016bidirectional} used by \newcite{levy2017zero} on their dataset. 

\begin{figure*}[ht!]
    \centering
    \begin{subfigure}[t]{0.49\textwidth}
        \centering
        \includegraphics[width=\textwidth]{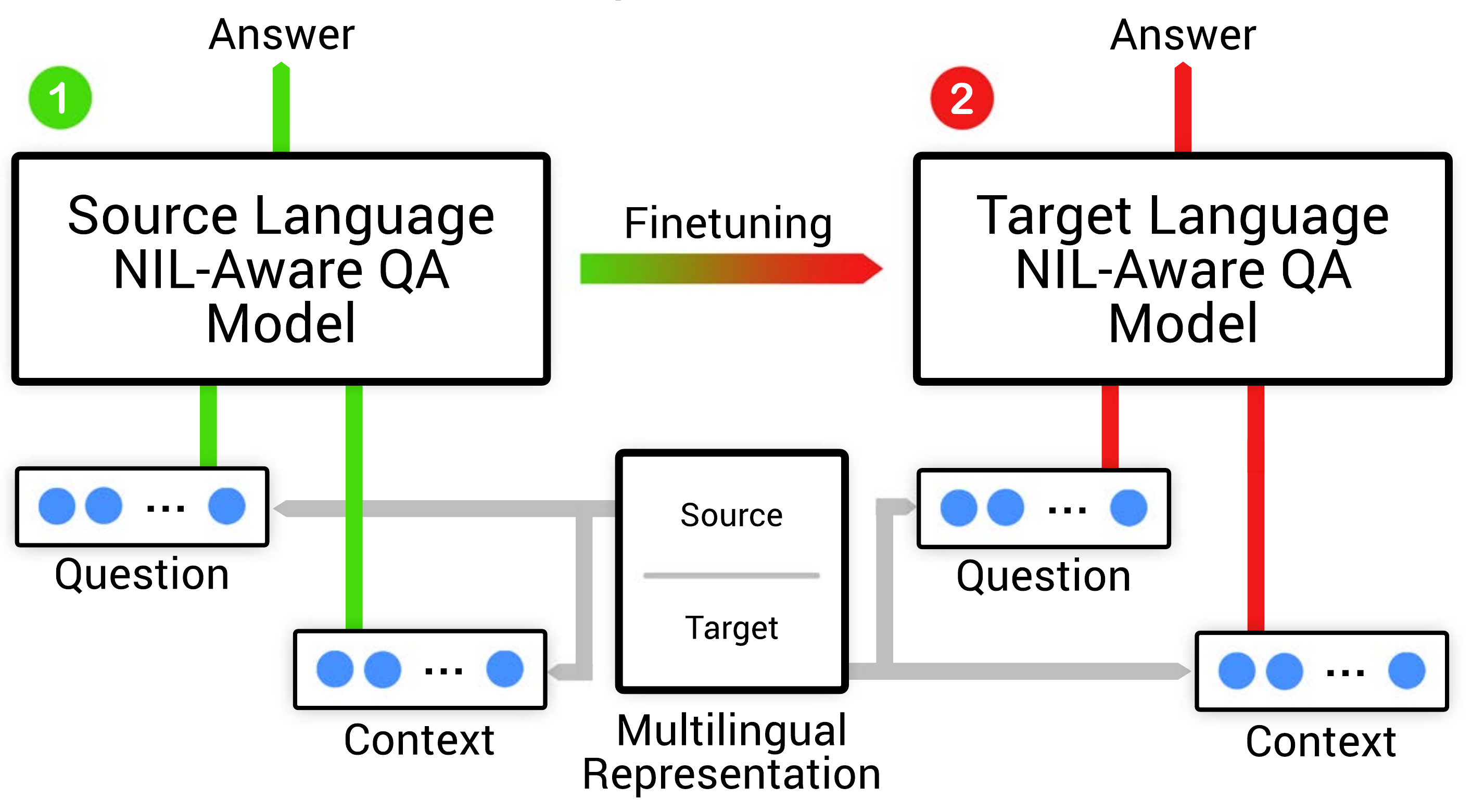}
        \caption{Cross-lingual model transfer. In step (1), a source language model is trained until convergence. In step (2), it is finetuned on a limited amount of target language data.}
        \label{subfig:setupTransfer}

    \end{subfigure}%
    ~ 
    \begin{subfigure}[t]{0.51\textwidth}
        \centering
        \includegraphics[width=\textwidth]{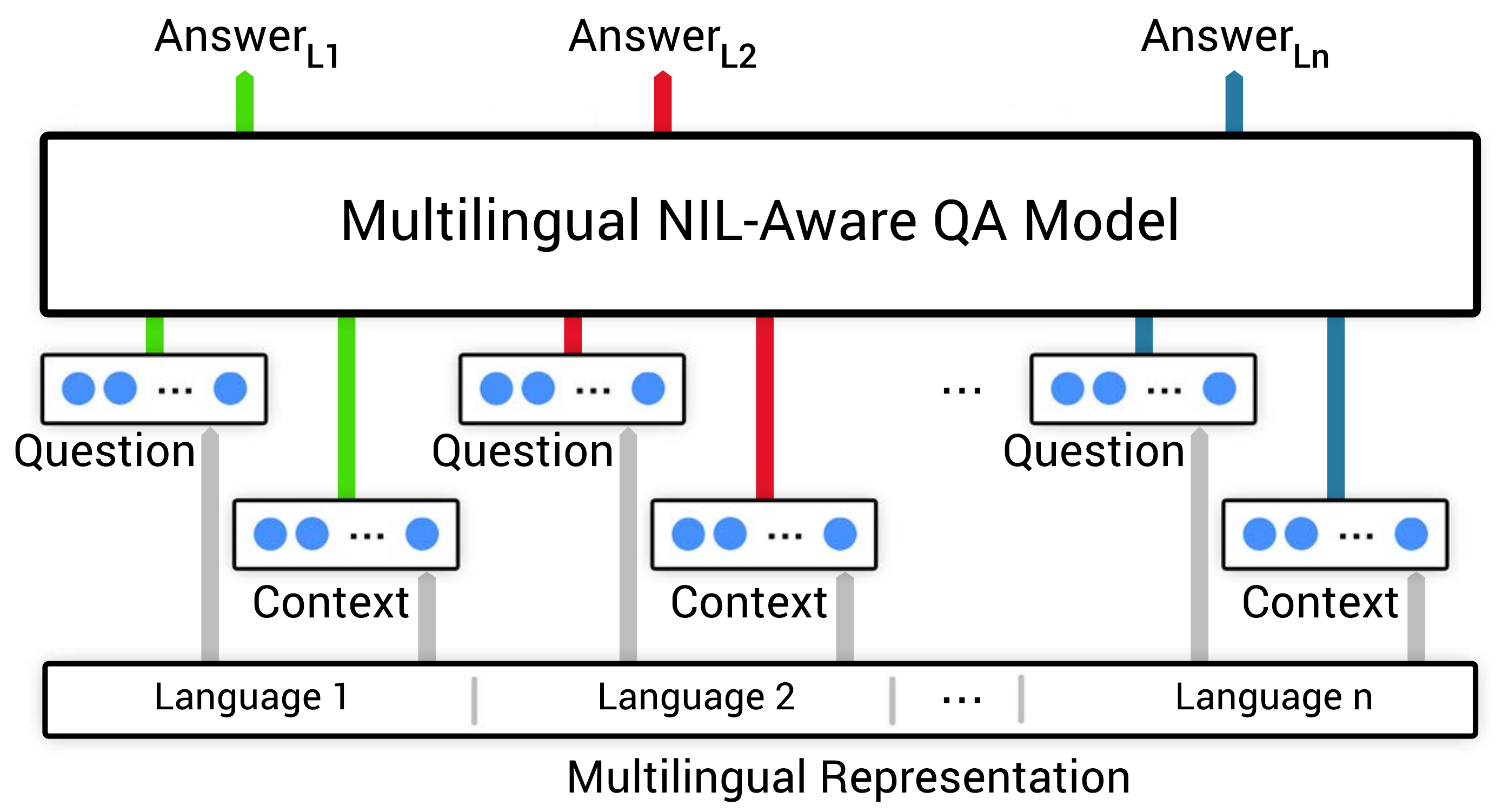}
        \caption{Joint multilingual training. }
        \label{subfig:setupMulti}

    \end{subfigure}
    \caption{Our cross-lingual transfer and multilingual training setups.}
\end{figure*}

\paragraph{A Nil-aware machine comprehension model} The reading comprehension model we employ, seen in Figure \ref{fig:Namanda}, encodes the question and context sequences and computes a similarity matrix between them. A column-wise softmax of the similarity matrix is multiplied with the question encoding to aggregate the most relevant parts of the question with respect to the context. Next, a joint-encoding of the question and context is created and a multi-factor self-attentive encoding is applied to accumulate evidence from the entire context. These representations are called the evidence vectors. Lastly, the evidence vectors are decomposed for every context word with orthogonal decomposition. The parallel components represent the relevant parts of the context and the orthogonal parts represent the irrelevant parts. These decompositions bias the decoder to either output a span or \texttt{NIL}.

\paragraph{Multilingual representations}
\label{subsec:multiling}
We compare two methods of obtaining multilingual representations. First, we employ fastText embeddings \citep{bojanowski2016enriching} mapped to a multilingual space in a supervised fashion \citep{conneau2017word}. Second, we employ the newly released multilingual BERT \citep{devlin2018bert} which is trained on the concatenation of the wikipedia corpora of 104 languages.\footnote{\url{https://github.com/google-research/bert/blob/master/multilingual.md}} For BERT, we take the contexualized word representations from the final layer as input to our machine comprehension model's question and context Bi-LSTM encoders. We do not fine-tune the pre-trained model.

\section{Experiments}
\label{expts}

Following \citet{levy2017zero}, 
we distinguish between the traditional RE setting where the aim is to generalize to unseen entities (\textbf{UnENT}) and the zero-shot setting (\textbf{UnREL}) where the aim is to do so for unseen relation types (see Section \ref{para:zero-shot}). Our goal is to answer these three questions: \textbf{A)} how well can RE models be transferred across languages? \textbf{B)} in the difficult \textbf{UnREL} setting, can the variance between languages in the number of instances of relations  (see Figure \ref{fig:2}) be exploited to enable more robust RE ? \textbf{C)} can one jointly-trained multilingual model which performs RE in multiple languages perform comparably to or outperform its individual monolingual counterparts? For all experiments, we take the \textit{multiple templates} approach where a model sees different paraphrases of the same question during training. This approach was shown by \newcite{levy2017zero} to have significantly better paraphrasing abilities than when only one question template or simpler relation descriptions are employed.

\paragraph{Evaluation} Our evaluation methodology follows \citet{levy2017zero}. We compute precision, recall and F1 by comparing spans predicted by the models with gold answers. Precision is equal to the
true positives divided by total number of non-nil answers predicted by a system. Recall is equal to the true positives divided by the total number of instances that are non-nil in the ground truth answers. Word order and punctuation are not considered.\footnote{We do not exclude articles from the evaluation as separating them from entities is not as trivial for other languages as it is for English.}

\subsection{Monolingual Baselines}
A baseline model is trained on the full monolingual training set (1 million instances) for each of the languages in both the \textbf{UnENT} and \textbf{UnREL} settings, which serve as a point of comparison for the cross-lingual transfer and multilingual models.

\paragraph{Comparison with \newcite{levy2017zero}}
\label{subsec:comparisonlevy}
In Table \ref{table:bigtable}, the comparison between the nil-aware machine comprehension framework we employ (Mono) and the results reported by \citet{levy2017zero} using the bias-augmented BiDAF model on their dataset (and splits) can be seen. The clear improvements obtained are in line with those reported by \newcite{kundu2018namanda} of NAMANDA over BiDAF on reading comprehension tasks.

\begin{figure*}[t!]
\centering
\includegraphics[width=\textwidth]{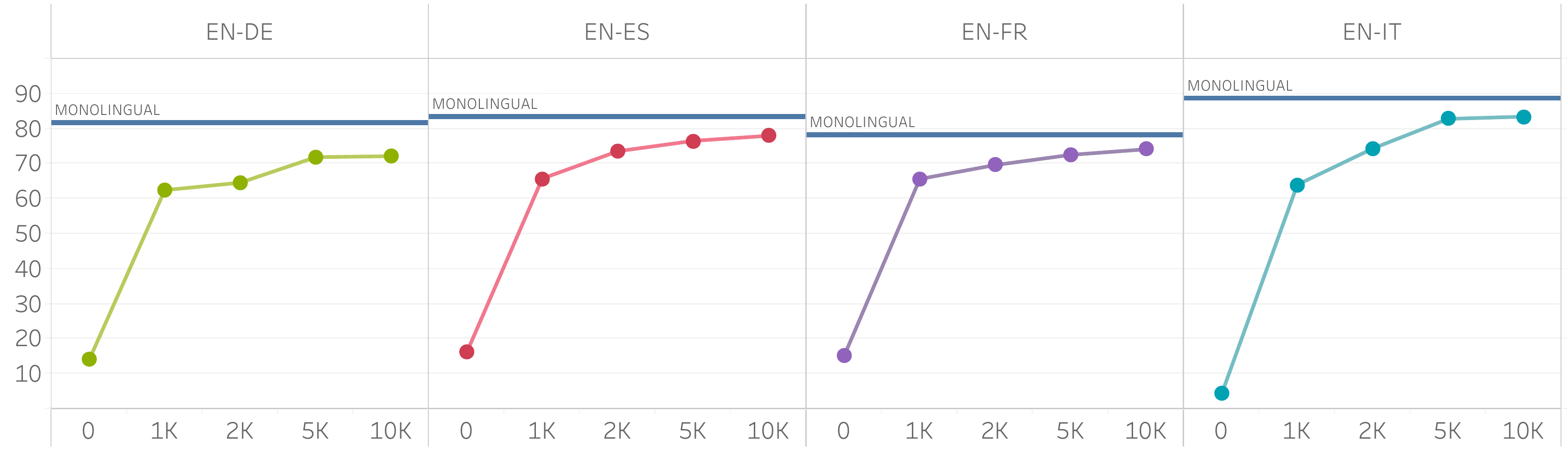}
\caption{F1-scores for the cross-lingual transfer experiments in the \textbf{UnENT} setting. The \textsc{Monolingual} line shows the corresponding monolingual model's F1-score. }
\label{fig:UnEntF1}
\end{figure*} 

\paragraph{Results}
Table \ref{table:bigtable} shows the results of the monolingual baselines. For the cross-lingual transfer experiments, these results can be viewed as a performance ceiling. 
 
Observe that the results on our dataset are in general lower than those reported in \newcite{levy2017zero}. This can be attributed to three factors: a) on average, the context length in our dataset is longer compared to  theirs (see Appendix \ref{appendix:vocabcoverage}); b) the fastText word embeddings we employ to facilitate multilingual sharing have a lower coverage of the vocabularies of each language than the GloVe word embeddings employed in that work; c) in the \textbf{UnREL} setting, we employ a more challenging setup of 5-fold cross-validation (as opposed to 10-fold in their experiments), meaning that a lower number of relations is seen at training time and the test set contains a higher number of unseen relations.

\subsection{Cross-Lingual Model Transfer} 
In this set of experiments, seen in Figure \ref{subfig:setupTransfer}, we test how well RE models can be transferred from a source language with a large number of training examples to target languages with no or minimal training data. In the \textbf{UnENT} experiments, we construct pairwise parallel test and development sets between English and each of the languages. An English RE model (built on top of the multilingual representations described in sub-section \ref{subsec:multiling}) is trained on a full English training set (1 million instances). We then evaluate how well this model can transfer to each of the four other languages in the following cases: with no finetuning or when 1000, 2000, 5000 or 10000 target language training examples are used for finetuning. Note that entities in the target languages' test and development sets are not seen in the English training data. We compare transfer performance with monolingual performance when a target language's full training set is employed.

A similar approach is followed for \textbf{UnREL} experiments. However, since the number of relations is relatively small, cross-validation with five folds is employed instead of fixed splits. Moreover, because this is a substantially more challenging setting we are interested in evaluating along another dimension (Question \textbf{B}): when relations are seen in the source language but not in the target language. Furthermore, unlike for \textbf{UnENT}, we directly use 10k examples for finetuning. 

\paragraph{Results} Figure \ref{fig:UnEntF1} shows the results of the cross-lingual transfer experiments for \textbf{UnENT}, where  transfer is accomplished through multilingually aligned fastText embeddings. In a parallel set of experiments, transfer was performed through the multilingual BERT encoder. The results of this (see Appendix \ref{appendix:results}) showed a clear advantage for the former over the latter.\footnote{We therefore continue the rest of our experiments in the paper using the multilingual fastText embeddings.} This is primarily due to the low vocabulary coverage of multilingual BERT which has a total vocabulary size of 100k tokens for 104 languages (see Appendix \ref{appendix:vocabcoverage} for coverage statistics). While it is clear that the models suffer from rather low recall when no finetuning is performed, the results show considerable improvements when finetuning with only 1000 target language examples. With 10K target language examples, it is possible to nearly match the performance of a model trained on the full target language monolingual training set. 

\begin{figure}[t]
\centering
\includegraphics[width=\columnwidth]{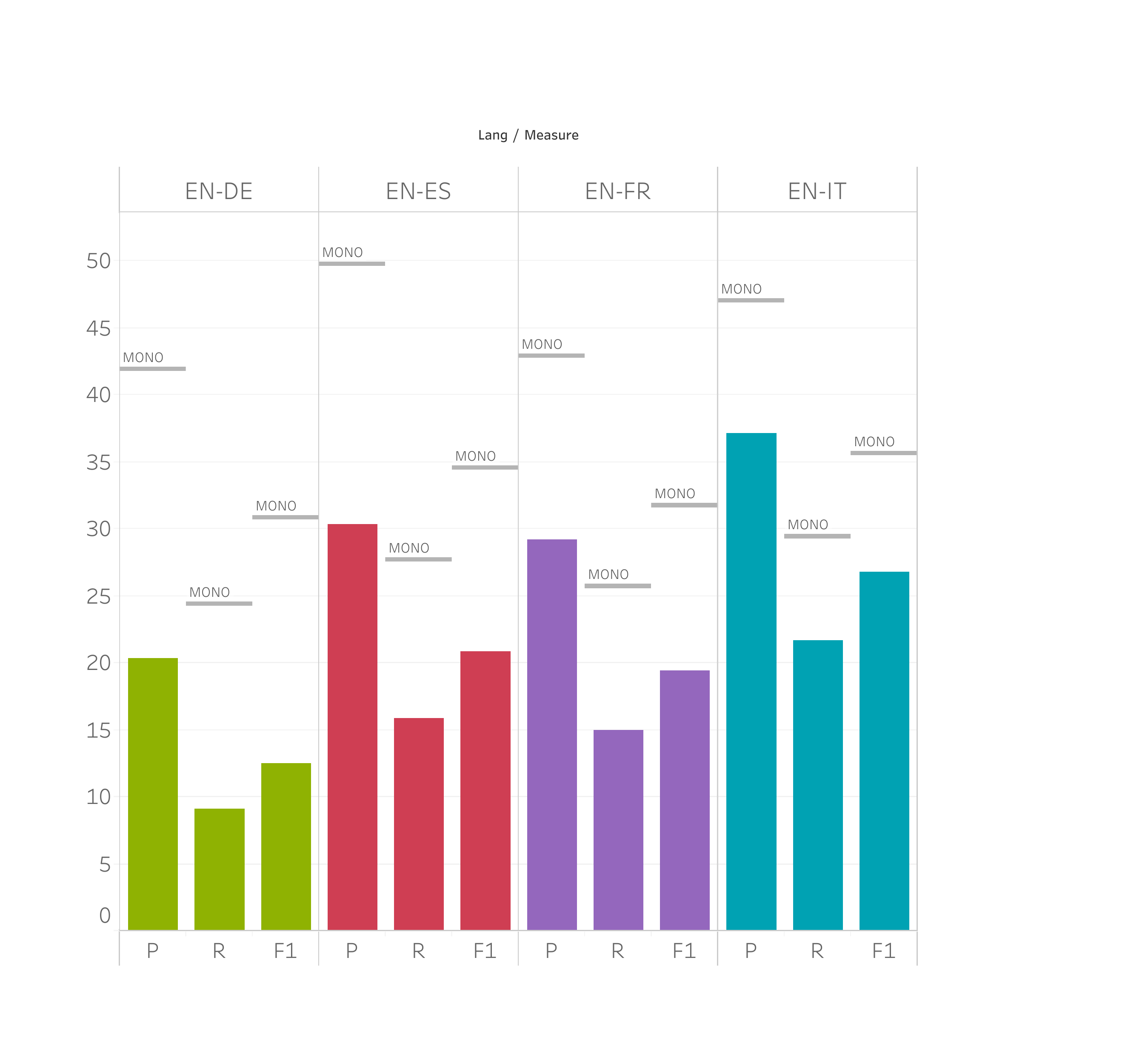}
\caption{Precision, Recall and F1-scores for the cross-lingual transfer experiments in \textbf{UnREL} setting. The results are the mean of 5-fold cross-validation. The \textsc{Mono} line shows the corresponding monolingual model's F1-score.}
\label{fig:UnREL}
\end{figure}

Similarly, in the \textbf{UnREL} experiments, our results (Figure \ref{fig:UnREL}) show that it's possible to recover a large part of the fully-supervised monolingual models' performance. It can be seen, however, that with 10k target language examples, a lower proportion of the performance is recovered when compared to the \textbf{UnENT} setting. This indicates that it is more difficult to transfer the ability to identify \textit{relation paraphrases} and \textit{entity types} through global cues\footnote{When context phrasing deviates from the question in a way that is common between relations.} which \newcite{levy2017zero} suggested are important for generalizing to new relations in this framework. 

\subsection{One Model, Multiple Languages}
We now examine the possibility of training one multilingual model which is able to perform relation extraction across multiple languages, as shown in Figure \ref{subfig:setupMulti}. We are interested in the case when an entity may be seen in another language's training data, as this is a realistic cross-lingual KB completion scenario where different languages' KBs are better populated for different topics. To control for training set size we include 200k training instances per language, so that the total size of the training set is equal to that of the monolingual baseline. However, an additional benefit of multilingual training is that extra overall training data becomes available. To test the effect of that we also run an experiment where the full training set of each of the languages is employed (adding up to 5 million training examples).

In the \textbf{UnREL} experiments, 5-fold cross-validation is performed. We are once again interested in exploiting the fact that KBs are better populated for different properties across different languages. Our setup is therefore as follows: in each of the 5 folds, a test set relation for a particular language is not seen in that language's training set, but may be seen in any of the other languages. This amounts to maintaining the original zero-shot setting (where a relation is not seen) monolingually, but providing supervision by allowing the models to \textit{peek across languages}.

\paragraph{Results} 
In the \textbf{UnENT} setting the multilingual models trained on just 200k instances per language perform slightly below the monolingual baselines. This excludes for French where, surprisingly, the baseline performance is actually exceeded. When the full training sets of all languages are combined, the multilingual model outperforms the monolingual baselines for three (English, Spanish, and French) out of five languages and is slightly worse for two (German and Italian). This demonstrates that not only is it possible to utilize a single model to perform RE in multiple languages, but that the multilingual supervision signal will often lead to improvements in performance. These results are shown in the third and fourth columns of Table \ref{table:bigtable}. 

The multilingual \textbf{UnREL} model outperforms its monolingual counterparts by large margins for all languages reaching a near 100\% F1-score improvement for most languages. 
This is largely in line with our premise that the natural topicality of KBs across languages can be exploited to provide cross-lingual supervision for relation extraction models.

\begin{table*}[ht!]
\fontsize{10}{10}\selectfont
  \centering
    \begin{tabular}{c|c|cccc|ccc}
    \toprule
    \multirow{2}{*}{Lang.} & &  \multicolumn{4}{c|}{UnENT} & \multicolumn{3}{c}{UnREL} \\

     &  &  \newcite{levy2017zero} & Mono. & Multi. (S) & Multi. (L) &  \newcite{levy2017zero} & Mono. & Multi. \\
     \midrule
     
    \multirow{3}{*}{EN*} & P  & 87.66 & \textbf{90.49} &   n/a   &   n/a   & 43.61 & \textbf{56.53} & n/a \\
                         & R  & 91.32 & \textbf{94.87} &   n/a   &   n/a   & 36.45 & \textbf{44.74} & n/a \\
                         & F1 & 89.44 & \textbf{92.63} &   n/a   &   n/a   & 39.61 & \textbf{49.85} & n/a \\

     \midrule
     \multirow{3}{*}{EN} & P  &   n/a   & 74.09 & 74.33 & \textbf{77.11} &   n/a   & 46.75 & \textbf{63.29} \\
                         & R  &   n/a   & 85.35 & 83.63 & \textbf{86.42} &   n/a   & 25.32 & \textbf{44.40} \\
                         & F1 &   n/a   & 79.32 & 78.71 & \textbf{81.50} &   n/a   & 32.78 & \textbf{51.99} \\

     \midrule
     \multirow{3}{*}{ES} & P  &   n/a   & 81.79 & 80.60 & \textbf{83.68} &   n/a   & 49.77 & \textbf{73.43} \\
                         & R  &   n/a   & \textbf{85.02} & 81.47 & 83.58 &   n/a   & 27.69 & \textbf{62.82} \\
                         & F1 &   n/a   & 83.37 & 81.03 & \textbf{83.63} &   n/a   & 34.54 & \textbf{67.64} \\

     \midrule
     \multirow{3}{*}{IT} & P  &   n/a   & \textbf{88.69} & 86.23 & 88.43 &   n/a   & 47.09 & \textbf{68.66} \\
                         & R  &   n/a   & \textbf{88.10} & 85.64 & 86.91 &   n/a   & 29.45 & \textbf{55.24} \\
                         & F1 &   n/a   & \textbf{88.39} & 85.93 & 87.66 &   n/a   & 35.62 & \textbf{61.13} \\

     \midrule
     \multirow{3}{*}{FR} & P  &   n/a   & 82.36 & 80.82 & \textbf{82.90} &   n/a   & 42.93 & \textbf{60.78} \\
                         & R  &   n/a   & 74.16 & 76.60 & \textbf{78.10} &   n/a   & 25.73 & \textbf{47.09} \\
                         & F1 &   n/a   & 78.05 & 78.66 & \textbf{80.43} &   n/a   & 31.78 & \textbf{53.06} \\

     \midrule
     \multirow{3}{*}{DE} & P  &   n/a   & \textbf{75.85} & 69.88 & 73.67 &   n/a   & 41.94 & \textbf{43.36} \\
                         & R  &   n/a   & \textbf{88.21} & 81.36 & 84.08 &   n/a   & 24.38 & \textbf{25.32} \\
                         & F1 &   n/a   & \textbf{81.57} & 75.20 & 78.53 &   n/a   & 30.82 & \textbf{31.97} \\

    \bottomrule
    \end{tabular}
      \caption{Precision, Recall, and F1-score results for all languages' monolingual (Mono.) and multilingual (Multi.) models. (S) indicates the small multilingual model which was trained on 200k examples and (L) indicates the large on trained on 5 million examples. * is used to mark the results on \newcite{levy2017zero}'s English dataset.}
  \label{table:bigtable}
\end{table*}

\subsection{Hyperparameters}
In all experiments, models were trained for five epochs with a learning rate of 1.0 using Adam \citep{kingma2014adam}. For finetuning in the cross-lingual transfer experiments, the learning rate was lowered to 0.001 to prevent forgetting and a maximum of 30 finetuning iterations over the small target language training set were performed with model selection using the target language development set F1-score. All monolingual models' word embeddings were initialised using fastText embeddings trained on each language's Wikipedia and common crawl corpora,\footnote{https://fasttext.cc/docs/en/crawl-vectors.html} except for the comparison experiments described in sub-section \ref{subsec:comparisonlevy} where GloVe \citep{pennington2014glove} was used for comparability with \newcite{levy2017zero}.

\section{Related Work}
\paragraph{Multilingual NLU}
Advances in natural language understanding tasks have been as impressive as they have been fast-paced. Until recently, however, the multilingual aspect of such tasks has not received as much attention. This is primarily due to the costs associated with annotating data for multiple languages. Recent work such as \newcite{conneau2018xnli, agic2017baselines} offer important benchmarks for evaluating cross-lingual transfer of natural language inference models. Similarly, \newcite{cer2017semeval} present a Semantic Textual Similarity dataset for four languages.

\paragraph{Multilingual relation extraction} Previous investigations of multilingual RE have been few and far between. \newcite{faruqui2015multilingual} employed a pipeline of machine translation systems to translate to English, then Open RE systems to perform RE on the translated text, followed by cross-lingual projection back to source language. \newcite{verga2015multilingual} apply the universal schema framework \citep{riedel2013relation} on top of multilingual embeddings to extract relations from Spanish text without using Spanish training data. This approach, however, only enables generalization to unseen entities and does not have the flexibility to predict unseen relations. Furthermore, both of these works faced a fundamental difficulty with evaluation. The former resort to 
manual annotation of a small number of examples (1000) in each language and the latter use the 2012 TAC Spanish slot-filling evaluation dataset in which ``\textit{the coverage
of facts in the available annotation is very small}''. With the introduction of \textbf{X-WikiRE}, this work provides the first large-scale dataset and benchmark for the evaluation of multilingual RE spanning five languages. While this paves the way for a wide range of research on multilingual relation extraction and knowledge base population, we hope to extend this to a larger variety of languages in future work, particularly as we have been able to show that the amount of training data required for cross-lingual model transfer is minimal, meaning that a small dataset (when only that is available) can go a long way.

\section{Conclusion}
We introduced \textbf{X-WikiRE}, a new, large-scale multilingual relation extraction dataset in which relation extraction is framed as a problem of reading comprehension to allow for generalization to unseen relations. Using this, we demonstrated that a) multilingual training can be employed to exploit the fact that KBs are better populated in different areas for different languages, providing a strong cross-lingual supervision signal which leads to considerably better zero-shot relation extraction; b) models can be transferred cross-lingually with a minimal amount of target language data for finetuning; c) better modelling of nil-awareness in reading comprehension models leads to improvements on the task. Our work is a step towards making KBs equally well-resourced across languages. To encourage future work in this direction, we release our code and dataset.

\bibliography{naaclhlt2019}
\bibliographystyle{acl_natbib}

\appendix

\newpage
\clearpage
\section{X-WikiRE descriptive statistics}
\label{appendix:multiwiki}
\begin{figure}[h!]
    \centering
    \includegraphics[width=0.32\textwidth]{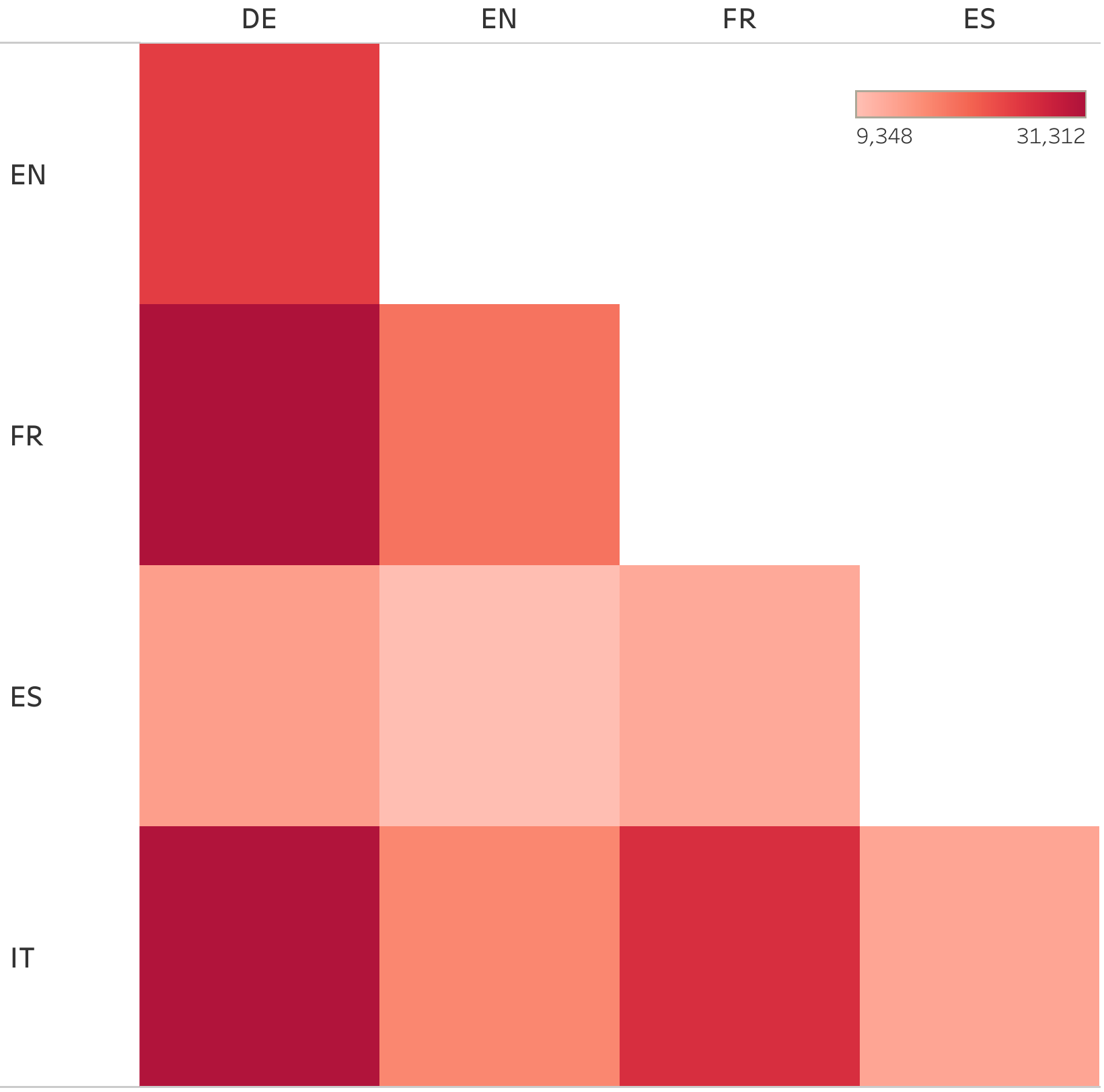}
    \caption{Property $place\_of\_death$.}
    \label{fig:intersection_single_p20}
\end{figure}

\begin{figure}[h!]
    \centering
    \includegraphics[width=0.32\textwidth]{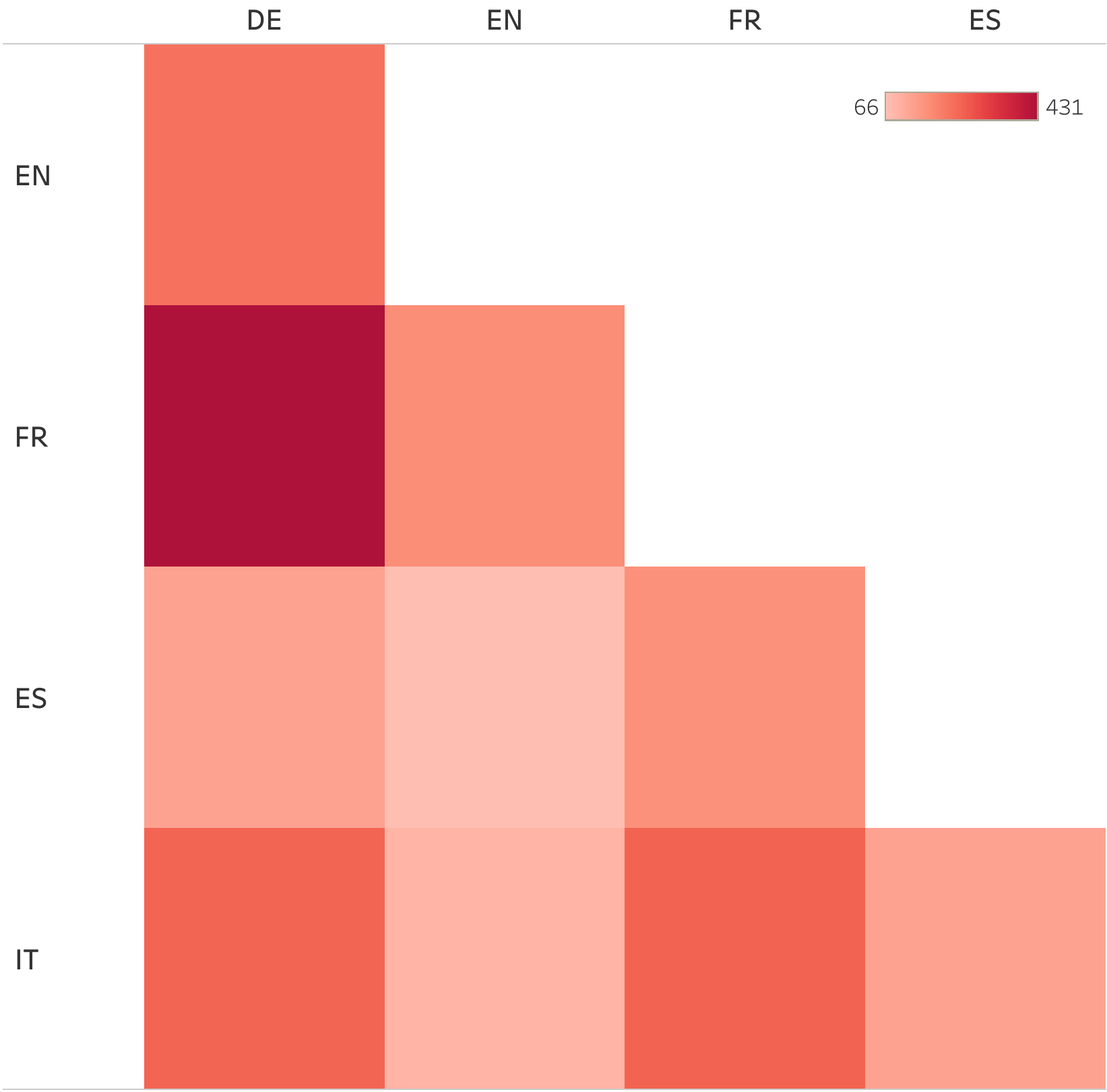}
    \caption{Property $military\_rank$.}
    \label{fig:intersection_single_p410}
\end{figure}

\begin{figure}[h!]
    \centering
    \includegraphics[width=0.32\textwidth]{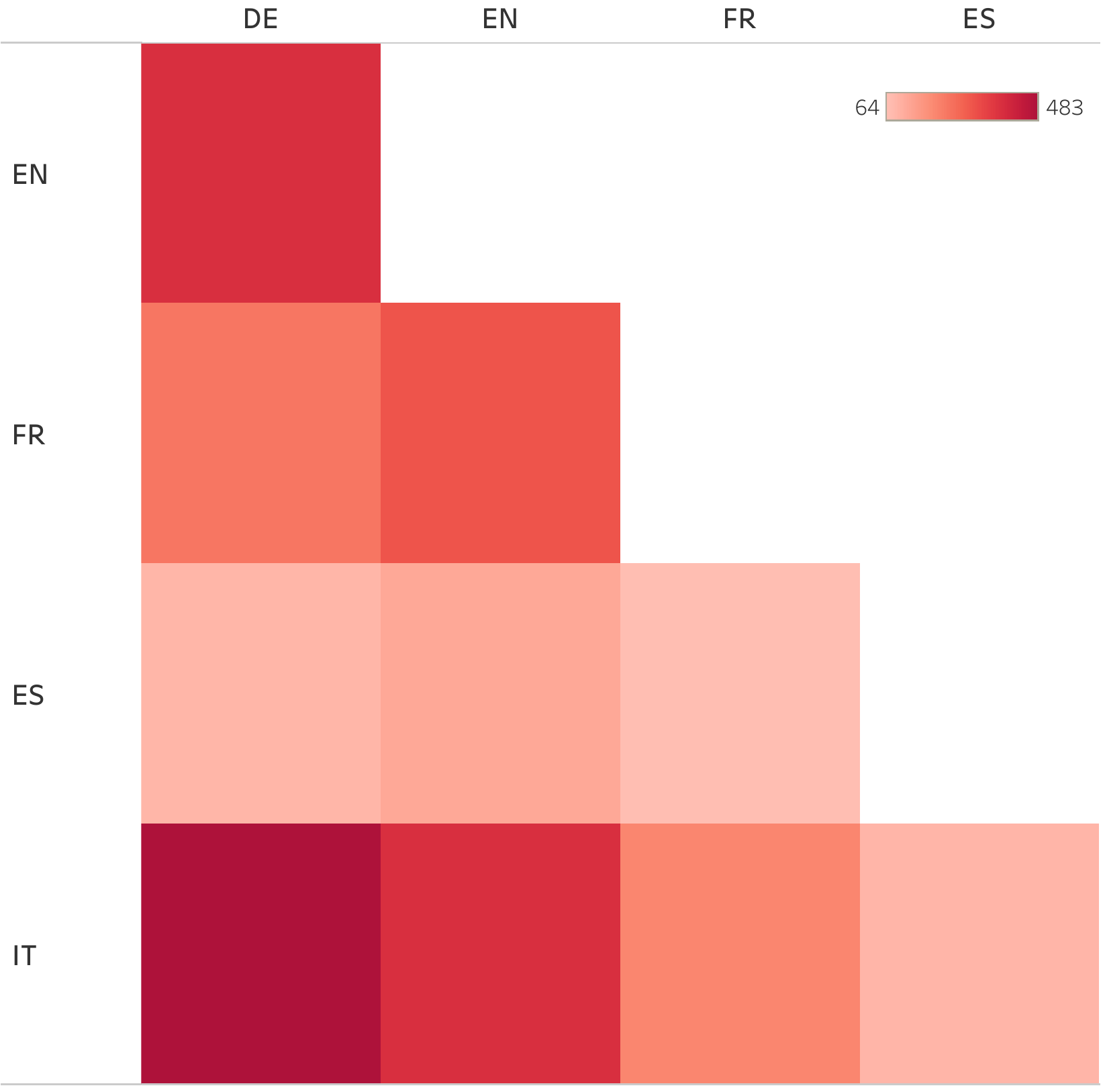}
    \caption{Property `$noble\_family$.}
    \label{fig:intersection_single_p53}
\end{figure}

\begin{figure}[h!]
    \centering
    \includegraphics[width=0.32\textwidth]{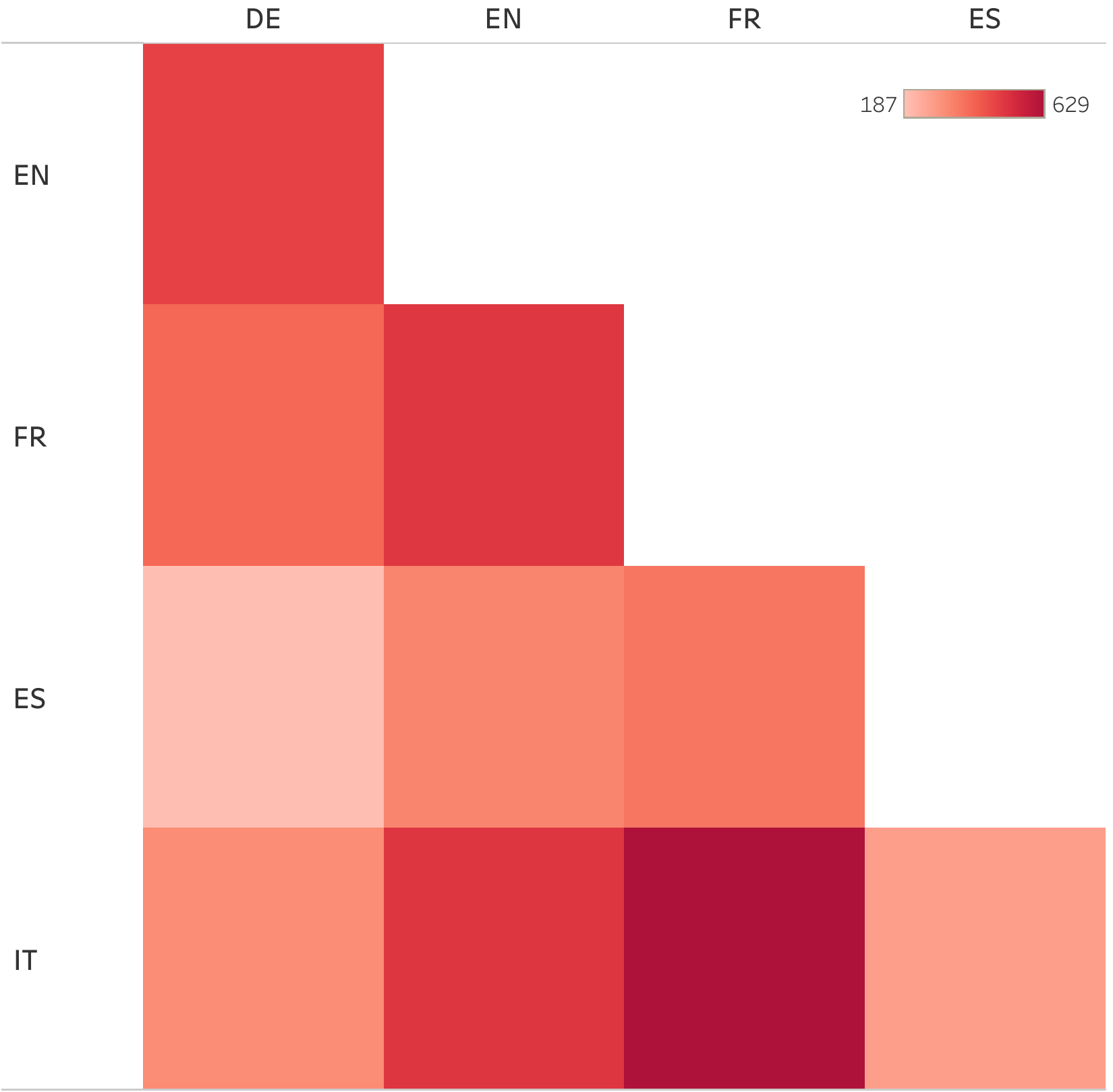}
    \caption{Property $start\_time$.}
    \label{fig:intersection_single_p580}
\end{figure}

 The distribution of Wikidata triples in Wikipedia text is not equal for each language. This also means that different languages share varying numbers of parallel triples per property. In Figures \ref{fig:intersection_single_p20}, \ref{fig:intersection_single_p410}, \ref{fig:intersection_single_p53}, and \ref{fig:intersection_single_p580} some examples are shown. 
    
\newpage
\section{Context size}
We computed the average length of the context in out dataset and \newcite{levy2017zero}'s dataset. Observe that in our dataset contexts are longer on average. Also observe that, on average, contexts in test set have more tokens than those in the training or development sets for all languages except Italian.

\begin{table}[h!]
  \centering
    \begin{tabular}{c|c|ccc}
    \toprule
    \multicolumn{1}{r}{} & Lang  & Train & Dev   & Test \\
    \midrule
    \newcite{levy2017zero} & EN*    & 29 & 28 & 29 \\
    \midrule
    \multirow{5}[2]{*}{X-WikiRE} & EN    & 30 & 33 & 34 \\
          & ES    & 34 & 46 & 48 \\
          & FR    & 49 & 39 & 42 \\
          & DE    & 29 & 30 & 32 \\
          & IT    & 40 & 35 & 35 \\
    \bottomrule
    \end{tabular}%
    \caption{Avarage number of tokens in the context.}
  \label{tab:addlabel}%
\end{table}%

\section{Vocabulary coverage}
\label{appendix:vocabcoverage}
 Table \ref{tab:lang_model_coverage} shows vocabulary coverage statistics for contexts and questions. Note that fastText has a higher coverage for all languages. 

\begin{table}[h!]
  \centering
  \begin{adjustbox}{width=\columnwidth}
    \begin{tabular}{c|cc|cc}
    \toprule
          & \multicolumn{2}{c|}{BERT} & \multicolumn{2}{c}{fastText} \\
    \midrule
    \multicolumn{1}{l|}{Lang} & Context & Question & Context & Question \\
    \midrule
    IT    & 24\%  & 35\%  & 64\%  & 71\%  \\
    FR    & 25\%  & 36\%  & 67\%  & 73\%  \\
    ES    & 24\%  & 37\%  & 65\%  & 73\%  \\
    DE    & 22\%  & 34\%  & 56\%  & 64\%  \\
    EN    & 30\%  & 37\%  & 63\%  & 71\%  \\
    \bottomrule
    \end{tabular}%
  \end{adjustbox}
  \caption{Vocabulary coverage for multilingual BERT and fastText.}
  \label{tab:lang_model_coverage}%

\end{table}%

\section{BERT vs fastText}
\label{appendix:results}

Table \ref{table:resultsunentcompare} shows the results for our model in the \textbf{UnENT} scenario using both multilingual BERT and fastText. BERT performs poorly compared to fastText in every language and almost for each of the finetuning settings. This is likely due to the lower coverage of our dataset's vocabulary as can be seen in Table \ref{tab:lang_model_coverage}.

\begin{center}

\begin{table}[h!]
\centering

\begin{adjustbox}{width=\columnwidth}

\begin{tabular}{c|r|c|c|c|c|c|c}
\toprule
\multirow{2}{*}{} & \multicolumn{1}{c|}{\multirow{2}{*}{Setting}} & \multicolumn{3}{c|}{BERT} & \multicolumn{3}{c}{fastText} \\ 
 & \multicolumn{1}{c|}{} & P & R & F1 & P & R & F1 \\ \hline
\multirow{5}{*}{\rotatebox[origin=c]{90}{EN-IT}} & 0 & 9.76 & 11.84 & \textbf{10.70} & 54.86 & 2.31 & 4.44 \\ 
 & 1K & 64.15 & 50.08 & 56.25 & 69.25 & 59.53 & \textbf{64.02} \\ 
 & 2K & 70.89 & 57.12 & 63.26 & 80.78 & 68.89 & \textbf{74.37} \\ 
 & 5K & 79.85 & 68.33 & 73.64 & 83.77 & 81.90 & \textbf{82.82} \\ 
 & 10K & 83.39 & 76.97 & 80.06 & 84.43 & 82.50 & \textbf{83.45} \\ \hline \hline 
\multirow{5}{*}{\rotatebox[origin=c]{90}{EN-ES}} & 0 & 10.22 & 3.75 & 5.49 & 28.98 & 11.22 & \textbf{16.17} \\ 
 & 1K & 55.19 & 35.19 & 42.97 & 67.56 & 64.08 & \textbf{65.78} \\ 
 & 2K & 68.43 & 50.36 & 58.02 & 74.45 & 72.73 & \textbf{73.58} \\ 
 & 5K & 72.65 & 63.97 & 68.04 & 79.23 & 74.03 & \textbf{76.54} \\ 
 & 10K & 79.76 & 65.06 & 71.66 & 79.27 & 76.76 & \textbf{77.99} \\ \hline \hline 
\multirow{5}{*}{\rotatebox[origin=c]{90}{EN-FR}} & 0 & 23.72 & 13.77 & \textbf{17.42} & 49.56 & 9.03 & 15.28 \\ 
 & 1K & 53.81 & 35.38 & 42.69 & 69.88 & 61.93 & \textbf{65.67} \\ 
 & 2K & 67.51 & 52.74 & 59.21 & 77.30 & 63.37 & \textbf{69.64} \\ 
 & 5K & 70.98 & 66.16 & 68.49 & 76.35 & 69.11 & \textbf{72.55} \\ 
 & 10K & 78.67 & 67.11 & 72.43 & 80.78 & 68.52 & \textbf{74.15} \\ \hline \hline 
\multirow{5}{*}{\rotatebox[origin=c]{90}{EN-DE}} & 0 & 3.33 & 2.52 & 2.87 & 21.00 & 10.60 & \textbf{14.09} \\ 
 & 1K & 50.84 & 62.35 & 56.01 & 55.84 & 70.91 & \textbf{62.47} \\ 
 & 2K & 58.46 & 65.96 & 61.99 & 56.67 & 75.13 & \textbf{64.60} \\ 
 & 5K & 61.37 & 71.75 & 66.16 & 68.00 & 76.24 & \textbf{71.88} \\ 
 & 10K & 66.65 & 74.66 & 70.43 & 66.89 & 78.37 & \textbf{72.17} \\ 
\bottomrule
\end{tabular}
\centering
\end{adjustbox}
\caption{Precision, Recall and F1-scores for \textbf{UnENT} comparing scores using BERT and fastText multilingual embeddings.}
\label{table:resultsunentcompare}

\end{table}
\end{center}

\end{document}